\documentclass{article} 
\usepackage{iclr2021_conference,times}

\usepackage{amsmath,amsfonts,bm}

\def\eqref#1{equation~\ref{#1}}

\def\1{\bm{1}}

\DeclareMathAlphabet{\mathsfit}{\encodingdefault}{\sfdefault}{m}{sl}
\SetMathAlphabet{\mathsfit}{bold}{\encodingdefault}{\sfdefault}{bx}{n}

\usepackage{hyperref}
\usepackage{url}
\usepackage{graphicx}
\usepackage{svg}
\usepackage{multirow}
\usepackage{booktabs}

\title{Online parameter inference for the simulation of a Bunsen flame using heteroscedastic Bayesian neural network ensembles}

\author{Maximilian L. Croci, Ushnish Sengupta \& Matthew P. Juniper\\
Department of Engineering\\
University of Cambridge\\
Cambridge, United Kingdom \\
\texttt{mlc70@cam.ac.uk} \\
}

\iclrfinalcopy 
\begin{document}

\maketitle

\begin{abstract}
This paper proposes a Bayesian data-driven machine learning method for the online inference of the parameters of a $G$-equation model of a ducted, premixed flame. Heteroscedastic Bayesian neural network ensembles are trained on a library of 1.7 million flame fronts simulated in LSGEN2D, a $G$-equation solver, to learn the Bayesian posterior distribution of the model parameters given observations. The ensembles are then used to infer the parameters of Bunsen flame experiments so that the dynamics of these can be simulated in LSGEN2D. This allows the surface area variation of the flame edge, a proxy for the heat release rate, to be calculated. The proposed method provides cheap and online parameter and uncertainty estimates matching results obtained with the ensemble Kalman filter, at less computational cost. This enables fast and reliable simulation of the combustion process.\footnote{Code available at: \url{https://github.com/nailimixaM/iclr-2021-baynne}}
\end{abstract}

\section{Introduction}
Simulation-based digital twins of physical systems are becoming cheaper to use due to improvements in computer processing power and storage~\citep{Fuller:2020}. Data-driven machine learning methods for the inference of digital twin model parameters from experiments are made possible in part due to the possibility of creating cheap synthetic data sets at scale. These methods can also address the need to do bring down the computational cost of parameter inference such that these models may be updated in real-time (online) using the latest sensor observations.

In the design of modern rocket engines and gas turbines, simulation of the combustion process (the flame) can be used to design out thermoacoustic instabilities which would otherwise lead to catastrophic damage~\citep{Juniper:2018}. Thermoacoustic instabilities are driven by the coupling of the heat release rate of the flame and the acoustics in the combustor~\citep{Rayleigh:1878}. It is therefore necessary to model the flame dynamics such that the heat release rate is quantitatively accurate. The $G$-equation~\citep{Williams:1985} is the kinematic model used in LSGEN2D~\citep{Hemchandra:2009} to simulate a ducted, premixed flame such as that from a Bunsen burner. By tuning the parameters of the $G$-equation model to fit a Bunsen flame experiment, a digital twin of the flame is created and its surface area variation in time, which is a proxy for the heat release rate, can be calculated. The ensemble Kalman filter (EnKF) is the current state-of-the-art, which iteratively performs Bayesian inference of the $G$-equation parameters from $G$-equation model forecasts generated in LSGEN2D and observations of the flame edge~\citep{Yu:2019}. The model forecasting is expensive, however, which makes the EnKF too expensive to be used online. We propose to infer online the $G$-equation parameters of Bunsen flame experiments using a heteroscedastic Bayesian neural network ensemble~\citep{Seng:2020, Seng:2020b}. In an expensive offline step, the ensemble learns a surrogate for the Bayesian posterior distribution of the parameters given observations of the flame front from a library of pairs of $G$-equation parameters and corresponding flame front shapes generated in LSGEN2D. The ensemble can then be used to infer online the parameters of the model for simulation of the Bunsen flames.

\section{Bunsen flame experiment and simulations}
\subsection{Bunsen flame experiment}
A Bunsen burner is placed inside a transparent duct and a high-speed camera is used to take images of the Bunsen flame at a frame rate of $f_s = 2500$ frames per second and a resolution of $1200\times800$ pixels. Speakers force the flame at frequencies in the range 250 Hz to 450 Hz. The gas composition (methane, ethene and air) and flow rate are varied using mass flow controllers. By varying the forcing frequency and amplitude and gas composition and flow rate, flames with different aspect ratios, propagation speeds and degrees of cusping of the flame front are observed. In some cases, the flame front cusping leads to pinch-off at the flame tip. For each of the 270 different flame operating conditions, 500 images are taken.

The flame images are processed to find a single-valued discretisation of the flame front, $x = f(y)$. First, the pixel intensities are thresholded and a position $x$ for every vertical position $y$ is found by weighted interpolation of the thresholded pixels, where the weights are the pixel intensities. Next, splines with 28 knots are used to smooth the $(x, y)$ coordinates. Each flame image is therefore converted into a $90\times1$ vector of flame front $x$ coordinates $\textbf{x}$ (as the $y$ coordinates are the same for all flames, they are discarded). Observation vectors $\textbf{z}$ are created by stacking 10 subsequent $\textbf{x}$ vectors. These observation vectors are used for inference with the neural networks. All 500 images of each Bunsen flame are processed in this way.

\subsection{Flame front model}
\begin{figure}[ht]
    \centering
    \includegraphics[width=13cm]{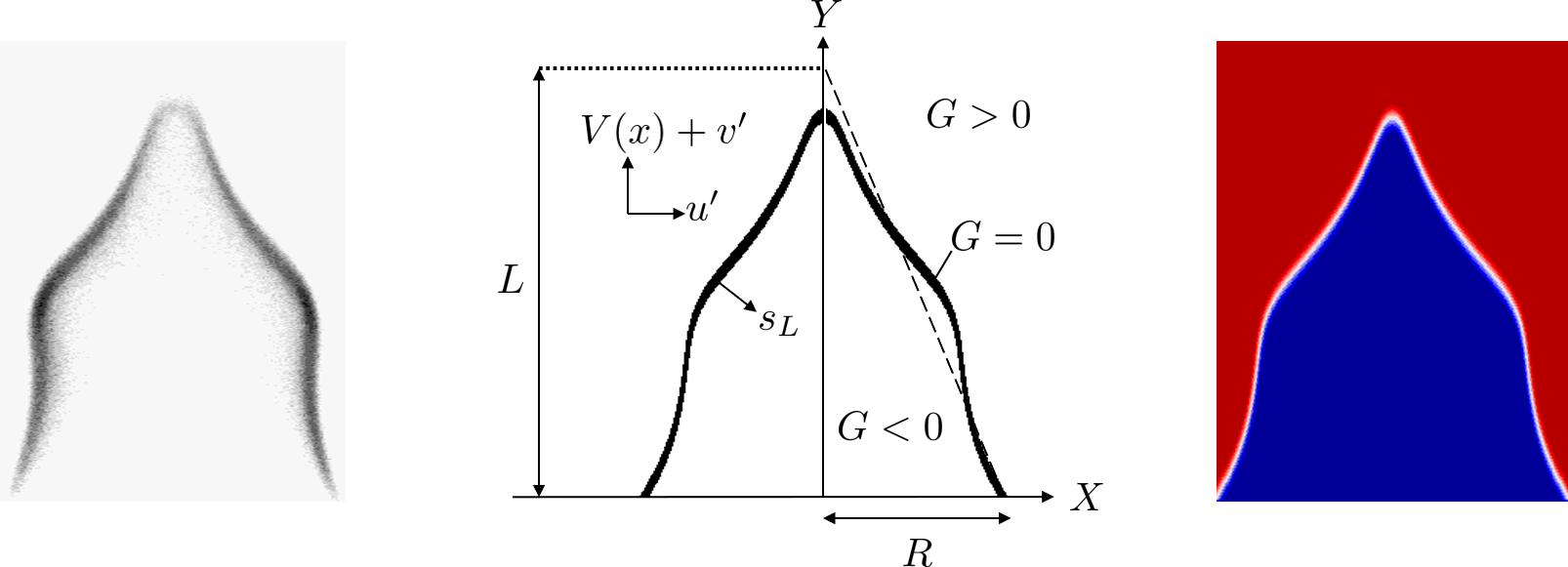}
    \caption{(\textit{Left}) Image of a Bunsen flame. (\textit{Middle}) $G$-equation model of the flame front. (\textit{Right}) Simulated flame in LSGEN2D.}\label{fig:flames}
\end{figure}

In this paper we use a kinematic model of the flame front as a boundary between reactants and products (see Figure \ref{fig:flames}). The flame front is defined to be the $G=0$ contour of a scalar field $G(x, y, t)$. Regions of negative and positive $G$ correspond to unburnt and burnt gases respectively (the magnitude of $G$ does not have a useful meaning). The position of the flame front in space and time is governed by:
\begin{equation}\label{eq:Geq}
    \frac{\partial G}{\partial t} + \textbf{v} \cdot \nabla G = s_L |\nabla G|,
\end{equation}
where $\textbf{v}$ is a prescribed velocity field and $s_L$ is the laminar flame speed: the speed at which the flame front propagates normal to itself into the reactants. The flame speed $s_L = s_L^0 \left( 1 - \mathcal{L}\kappa \right)$ is a function of the unstretched (adiabatic) flame speed $s_L^0$, the flame curvature $\kappa$ and the Markstein length $\mathcal{L}$, and is insensitive to pressure variations.
The unstretched flame speed $s_L^0$ depends only on the flame chemistry. The velocity field $\textbf{v} = u'\textbf{i} + \left( V(x) + v' \right)\textbf{j}$ comprises a parabolic base flow profile $V(x) = V(1 + \alpha(1 - 2(\frac{x}{R})^2))$ and superimposed continuity-obeying velocity perturbations $u'(x, y, t) = V\epsilon \sin (\text{St}(\frac{K}{R}y - t))$ and $v'(x, y, t) =  -\frac{V\epsilon K\text{St}}{R}x\cos(\text{St}(\frac{K}{R}y - t))$
where $\alpha$ determines the shape of the base flow profile ($\alpha = 0$ is uniform flow, $\alpha = 1$ is Poiseuille flow), $\epsilon$ is the amplitude of the vertical velocity perturbation with phase speed $V/K$, $\text{St} = 2\pi f R \beta/V$ is the Strouhal number with forcing frequency $f$ and flame radius $R$, and $\beta$ is the aspect ratio of the unperturbed flame. The parameters $K, \epsilon, \mathcal{L}, \alpha, \text{St}$ and $\beta$ are tuned to fit an observed flame shape.

\subsection{Simulated flame front library}
A library of simulated flame fronts with known parameters $K, \epsilon, \mathcal{L}, \alpha, \text{St}$ and $\beta$ is created for neural network training. The parameter values are sampled using quasi-Monte Carlo sampling to ensure good coverage of the parameter space. The parameters are sampled from the following ranges: $0.0 < K \leq 1.5$, $0.0 < \epsilon \leq 1.0$, $0.02 \leq \mathcal{L} \leq 0.08$, $0.0 \leq \alpha \leq 1.0$, $2.0 \leq \beta \leq 10.0$ and $0.08 \leq f/f_s \leq 0.20$. The values of $\text{St}$ are calculated by additionally sampling $0.002 \leq R \leq 0.004$ m and $1 \leq V \leq 5$ m/s and calculating $\text{St} = 2\pi f R \beta/V$. The parameters are sampled 8500 times, normalised to between 0 and 1 and recorded in target vectors $\{\textbf{t}=[K, \epsilon, \mathcal{L}, \alpha, \text{St}, \beta]\}$.

For each of the 8500 unique parameter configurations, LSGEN2D iterates the $G$-equation model of the flame front until it converges to the corresponding forced cycle. For each of the 200 forced cycle $G$ field states produced by LSGEN2D, the flame front is found by interpolating the $G$ values. This results in a $x = f(y)$ discretisation and the vectors $\textbf{x}$ of $x$ coordinates are recorded. Observation vectors $\textbf{z}$ are created by stacking 10 subsequent $\textbf{x}$ vectors. There are 200 observation vectors created from every cycle, resulting in a library of $1.7\times 10^6$ observation-target parameter pairs $\{(\textbf{z}, \textbf{t})\}$. This library is split 80\%-20\% into training and testing data sets.

\section{Inference using heteroscedastic Bayesian neural network ensembles}
\begin{figure}[ht]
    \centering
    \includegraphics[width=11cm]{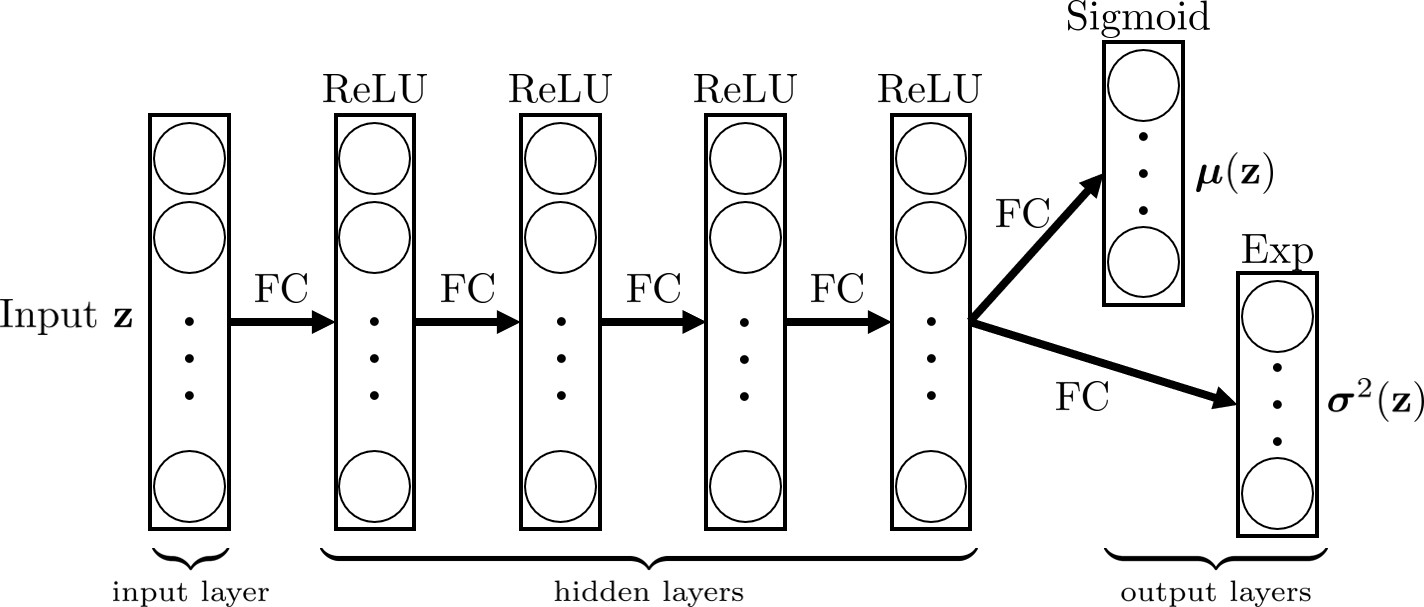}
    \caption{Architecture of a single neural network. The input and hidden layers have 900 nodes each, while the output layers have 6 nodes each. All layers are fully connected (FC). Rectified Linear Unit (ReLU) activation functions are used for the hidden layers and sigmoid and exponential (Exp) activation functions are used for the mean and variance ouput layers respectively.}\label{fig:NN}
\end{figure}

We assume the posterior probability distribution of the parameters given the observations can be modelled by a neural network: $p_\theta (\textbf{t} | \textbf{z})$ with its own parameters $\theta$. We assume this posterior distribution has the form:
\begin{equation}
    p_\theta (\textbf{t} | \textbf{z}) = N\left(\textbf{t}; \boldsymbol\mu(\textbf{z}), \boldsymbol\Sigma(\textbf{z})\right),
\end{equation}
where $\boldsymbol\Sigma(\textbf{z}) = \text{diag}(\boldsymbol\sigma^2(\textbf{z}))$. This encodes our assumption that the parameters are all mutually independent given the observations $\textbf{z}$. The architecture of a single neural network is shown in Figure \ref{fig:NN}. Each neural network comprises an input layer, four hidden layers with ReLU activations and two output layers: one for the mean vector $\bm{\mu}(\textbf{z})$ and one for the variance vector $\bm{\sigma}^2(\textbf{z})$. The output layer for the mean uses a sigmoid activation to restrict outputs to the range $(0, 1)$. The output layer for the variance uses an exponential activation to ensure positivity. An ensemble of $M = 20$ such neural networks is initialised, each with unique initial weights $\bm{\theta}_{j, anc}$ sampled from a Gaussian prior distributions according to He initialisation~\citep{He:2015}.

For a single observation $\textbf{z}$, the $j$-th neural network in the ensemble produces a sample of the posterior with an associated estimate of the aleatoric noise in the observations: $\boldsymbol\mu_j(\textbf{z}), \ \boldsymbol\sigma_j^2(\textbf{z})$.
This is achieved by using the loss function $\mathfrak{L}_j$:
\begin{equation} \label{eq:loss}
\mathfrak{L}_j = \left(\bm{\mu}_j(\textbf{z}) - \textbf{t}\right)^T \bm{\Sigma}_j\left(\textbf{z}\right)^{-1}\left(\bm{\mu}_j(\textbf{z}) - \textbf{t}\right) + \log\left(|\bm{\Sigma}_j\left(\textbf{z}\right)|\right)
 + \left(\bm{\theta}_j - \bm{\theta}_{anc, j}\right)^T\bm{\Sigma}_{prior}^{-1}\left(\bm{\theta}_j - \bm{\theta}_{anc, j}\right).
\end{equation}
The loss function comprises the negative log of the Gaussian likelihood function (probability of the observations given the targets, first two terms) and a regularising (penalty) term. By regularising about parameter values drawn from a prior distribution, the NNs produce samples from the posterior distribution. This is called randomised maximum a-posteriori (MAP) sampling~\citep{Pearce:2020}.

Once converged, the prediction from the ensemble for an observation $\textbf{z}$ is therefore a mixture of $M$ Gaussians each centered at their respective means $\boldsymbol\mu_j(\textbf{z})$. This mixture is approximated by a single multivariate Gaussian posterior distribution $p(\textbf{t}|\textbf{z}) \approx \mathcal{N}(\textbf{t}; \boldsymbol\mu(\textbf{z}),  \boldsymbol\Sigma(\textbf{z}))$ with mean $\boldsymbol\mu(\textbf{z}) = \frac{\Sigma_j  \boldsymbol\mu_j(\textbf{z})}{M}$ and covariance $\boldsymbol\Sigma(\textbf{z}) = \text{diag}(\boldsymbol\sigma^2(\textbf{z}))$ where $ \boldsymbol\sigma^2(\textbf{z}) = \frac{\Sigma_j \boldsymbol\sigma_j^2(\textbf{z})}{M} + \frac{\Sigma_j  \boldsymbol\mu_j^2(\textbf{z})}{M} - \left(\frac{\Sigma_j  \boldsymbol\mu_j(\textbf{z})}{M}\right)^2$ following similar treatment in~\cite{Lakshminarayanan:2017}. This is done for every observation vector $\textbf{z}$. The posterior distribution $p(\textbf{t}|\textbf{z}_i)$ with the smallest total variance $\sigma^2_{i, \text{tot}} = ||\boldsymbol\sigma^2(\textbf{z}_i)||_1$ is chosen as the best guess to the true posterior. The $M$ parameter samples from the chosen posterior are used for re-simulation, which allows us to check the predicted flame shapes and to calculate the normalised area variation over one cycle. 

\section{Results}
\begin{figure}
    \centering
    \includegraphics[width=12cm]{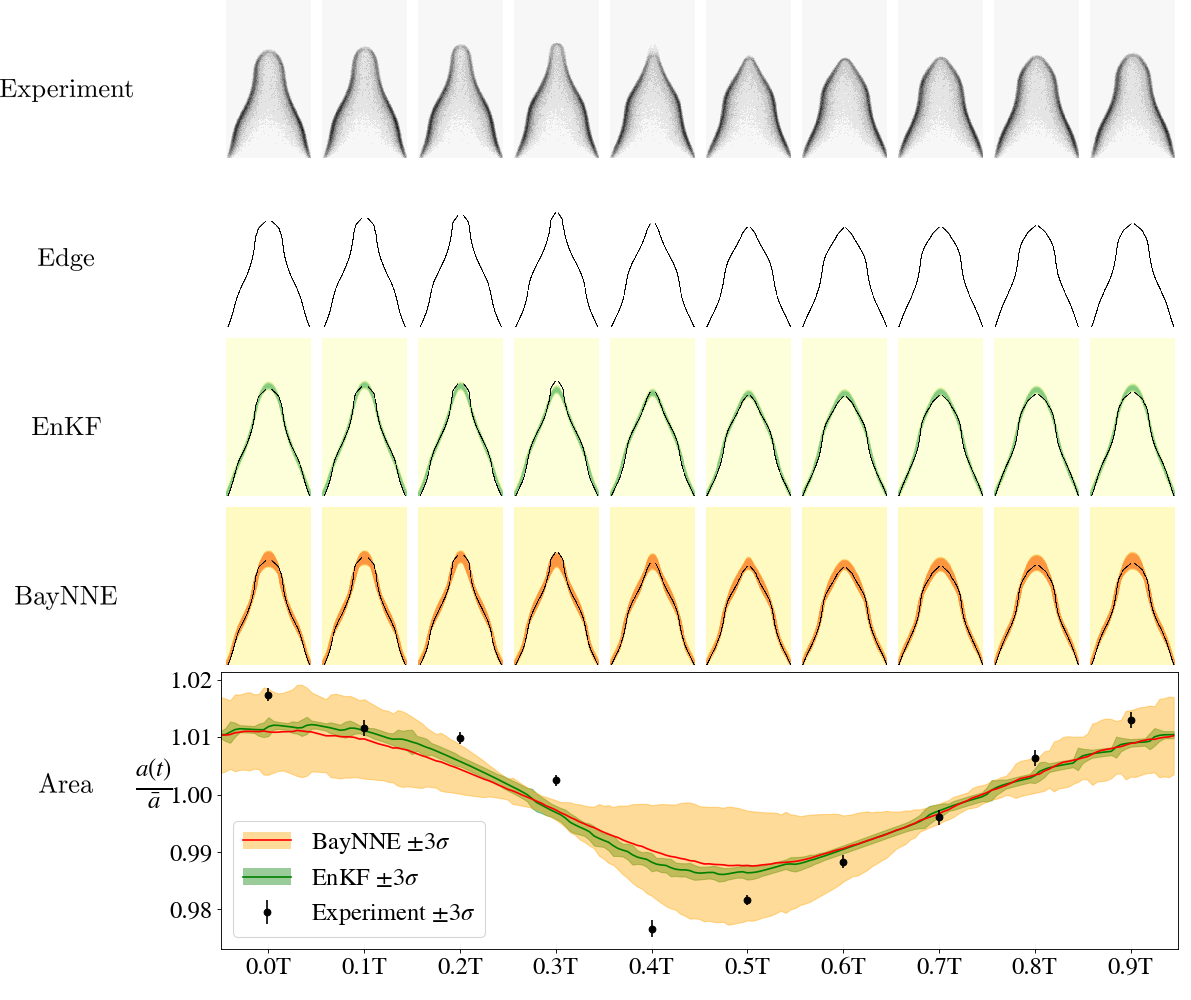}
    \caption{Results of inference using the BayNNE method compared to the EnKF method for the simulation of two different Bunsen flames. The flame images are preprocessed to find the flame front and then used for inference. The green and orange bands in the EnKF and BayNNE predicted flame shape plots are regions of high likelihood. The normalised area variations over one period of the simulated flames show good agreement with the experiments.
    }\label{fig:res2}
\end{figure}

The ensemble is trained for 5000 epochs and evaluated on the Bunsen flame image data. Parameter predictions for an observation vector $\textbf{z}$ take $O(10^{-4})$ seconds on an Nvidia P100 GPU. Figure \ref{fig:res2} show the results of inference and simulation of two different Bunsen flames. The BayNNE predicts flame shapes in good agreement with the experiments: the root mean square distances between the simulated and Bunsen flame shapes of 5 different Bunsen flames ranges from 0.017 units to 0.024 units, with mean 0.019 units (where the flame radius is 1 unit). These results show that the BayNNE's parameter estimates match those of the EnKF and require $O(10^8)$ less computing power.

The uncertainty in the BayNNE's predictions is greater than that of the EnKF due to the calculated posterior being the probability of the parameters given 10 flame images, whereas the EnKF considers all 500 flame images of the Bunsen flame. Unfortunately, it is not possible to combine the BayNNE's parameter estimates without knowledge of the probability distribution between the observation vectors. Any two observation vectors are not independent as knowledge about the first restricts our expectation of the second to a likely set of forced cycle states. Future work will address this limitation by using alternative neural network architectures, such as long-short term memory networks~\citep{Hochreiter:1997}, that are better suited to time-series data.

\section{Conclusions}
This work proposes a method for inferring the parameters of the $G$-equation model of a Bunsen flame for simulation purposes. Bayesian inference of the parameters and their uncertainties is performed using heteroscedastic Bayesian neural network ensembles. The neural networks are trained on a library of synthetic flame front observations created using LSGEN2D. Once trained, the Bayesian neural network ensemble accurately predicts the parameters and uncertainties from just 10 images of the Bunsen flame. The simulated flames with parameters predicted by the ensemble agree with the experiments, and the surface area variation, which is a proxy for the heat release rate, is calculated. A quantitatively accurate digital twin of the Bunsen flame is therefore created. Future work will focus on improving the parameter and uncertainty estimates by leveraging alternative neural network architectures.

\begin{figure}
    \centering
    \includegraphics[width=12cm]{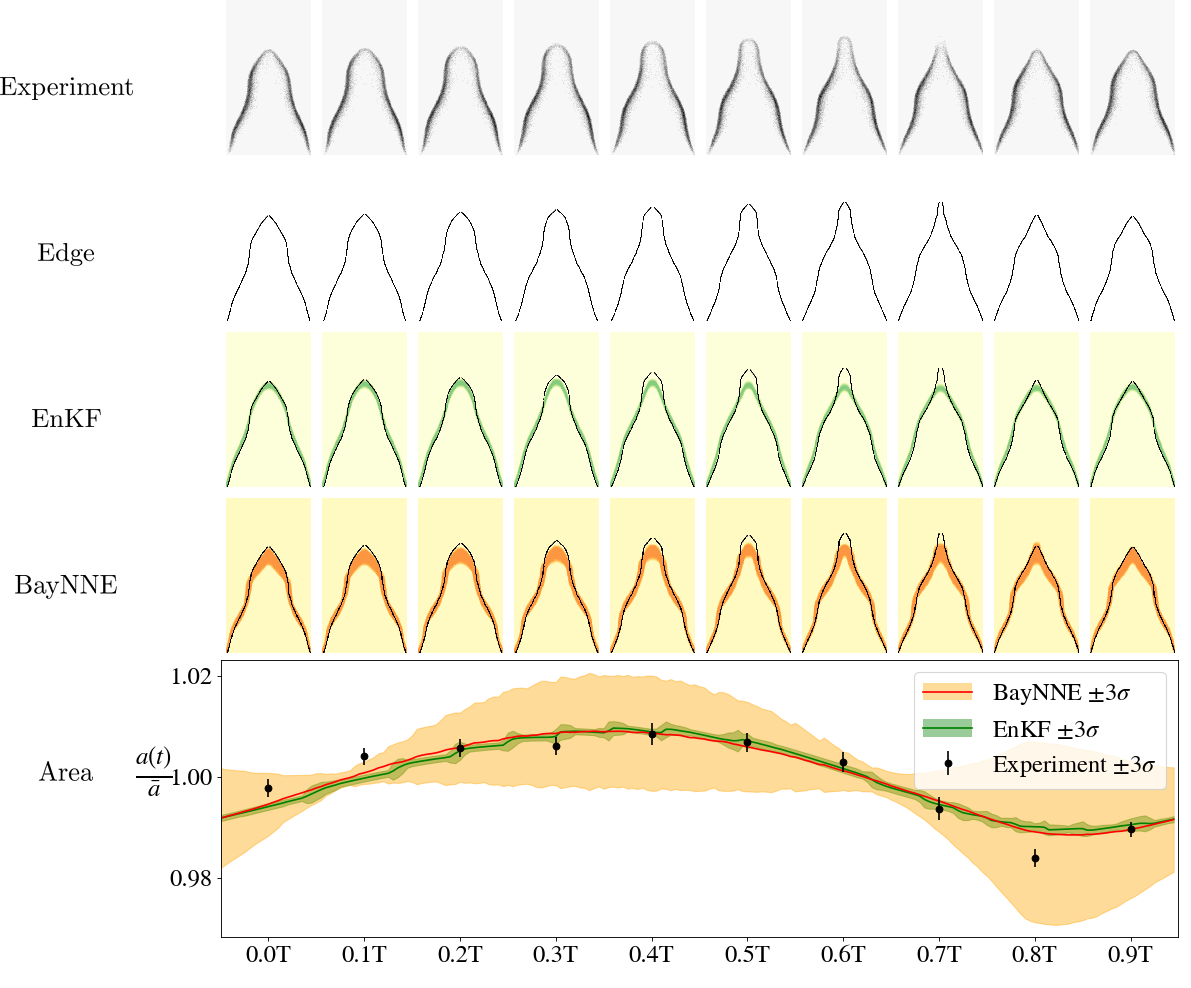}
\end{figure}
\section*{Disclosure of Funding}
This project has received funding from the UK Engineering and Physical Sciences Research Council (EPSRC) award EP/N509620/1 and from the European Union’s Horizon 2020 research and innovation program under the Marie Skłodowska-Curie grant agreement number 766264.

\bibliography{iclr2021_conference}
\bibliographystyle{iclr2021_conference}

\appendix
\section{Supplementary material: Hyperparameter settings}
\begin{table}[h]
\centering
\caption{Hyperparameter settings.}
\begin{tabular}{lr}
\toprule
Hyperparameter & Value  \\
\midrule
\textit{Training} & \\
\hspace{3mm}Train-test split & 80:20 \\
\hspace{3mm}Batch size & 2048 \\
\hspace{3mm}Epochs & 5000 \\
\hspace{3mm}Optimiser & Adam \\
\hspace{3mm}Learning rate & $10^{-3}$ \\
\midrule
\textit{Architecture} & \\
\hspace{3mm}Input units & 900 \\
\hspace{3mm}Hidden layers & 4 \\
\hspace{3mm}Units per hidden layer & 900 \\
\hspace{3mm}Output layers & 2 \\
\hspace{3mm}Units per output layer & 6  \\
\hspace{3mm}Ensemble size & 20 \\
\bottomrule
\end{tabular}
\label{tab:hyp}
\end{table}

\end{document}